\documentclass[11pt,a4paper]{article}

\usepackage[draft]{minted}
\usepackage{algorithm}
\usepackage[]{algpseudocode}
\algrenewcommand\algorithmicindent{0.76em}

\usepackage[hyperref]{acl2021}
\usepackage{times}
\usepackage{subfig}
\usepackage{amsmath}
\usepackage{latexsym}

\usepackage{graphicx}
\usepackage{booktabs}
\usepackage{makecell}
\usepackage{tabularx}
\usepackage{multirow}
\usepackage{setspace, enumitem,titlesec}

\graphicspath{{figs/}}
\DeclareGraphicsExtensions{.pdf,.png,.jpg}

\usepackage{microtype}

\aclfinalcopy 

\title{CommitBERT: Commit Message Generation \\
Using Pre-Trained Programming Language Model}

\author{
Tae-Hwan Jung \\
Kyung Hee University \\
\tt{nlkey2022@gmail.com}
}

\date{}

\begin{document}
\maketitle

\begin{abstract}
In version control using Git, the commit message is a document that summarizes source code changes in natural language.
A good commit message clearly shows the source code changes, so this enhances collaboration between developers.
To write a good commit message, the message should briefly summarize the source code changes, which takes a lot of time and effort.
Therefore, a lot of research has been studied to automatically generate a commit message when a code modification is given.
However, in most of the studies so far, there was no curated dataset for code modifications (additions and deletions) and corresponding commit messages in various programming languages.
The model also had difficulty learning the contextual representation between code modification and natural language.

To solve these problems, we propose the following two methods:
(1) We collect code modification and corresponding commit messages in Github for six languages (Python, PHP, Go, Java, JavaScript, and Ruby) and release a well-organized 345K pair dataset.
(2) In order to resolve the large gap in contextual representation between programming language (PL) and natural language (NL), we use CodeBERT \cite{feng2020codebert}, a pre-trained language model (PLM) for programming code, as an initial model.
Using two methods leads to successful results in the commit message generation task.
Also, this is the first research attempt in fine-tuning commit generation using various programming languages and code PLM.
Training code, dataset, and pretrained weights are available at \href{https://github.com/graykode/commit-autosuggestions}{https://github.com/graykode/commit-autosuggestions}.
\end{abstract}

\section{Introduction}
\label{sec:introduction}

\begin{figure}[t!]
    \centering
    \includegraphics[width=0.49\textwidth]{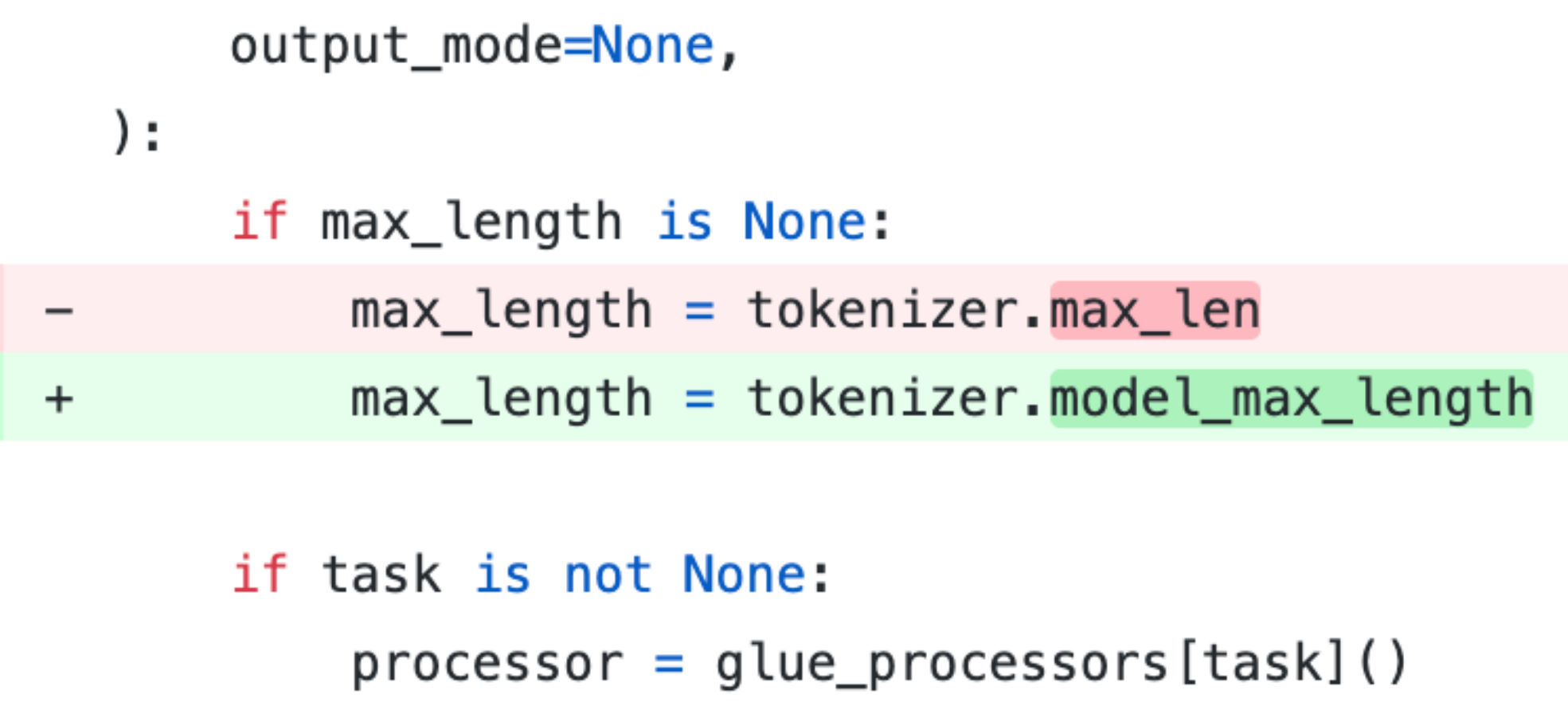}\\
    Message : fix deprecated ref to tokenizer.max\_len
    
    \caption{
    \label{fig:codediff}
    The figure above shows an example of commit message and git diff in Github.
    In the Git process, git diff uses unified format (unidiff \footnotemark):
    A line marked in red or green means a modified line, and green highlights in '+' lines are the added code, whereas red highlights in '-' lines are the deleted code.
    }
\end{figure}

Commit message is the smallest unit that summarizes source code changes in natural language.
Figure~\ref{fig:codediff} shows the git diff representing code modification and the corresponding commit message.
A good commit message allows developers to visualize the commit history at a glance, so many teams try to do high quality commits by creating rules for commit messages.
For example, Conventional Commits \footnote{\href{https://conventionalcommits.org}{https://conventionalcommits.org}} is one of the commit rules to use a verb of a specified type for the first word like 'Add' or 'Fix' and limit the length of the character.
It is very tricky to follow all these rules and write a good quality commit message, so many developers ignore it due to lack of time and motivation.
So it would be very efficient if the commit message is automatically written when a code modification is given.

\footnotetext{\href{https://en.wikipedia.org/wiki/Diff\#Unified_format}{https://en.wikipedia.org/wiki/Diff}}

Similar to text summarization, many studies have been conducted by taking code modification $X=(x_1,...,x_n)$ as encoder input and commit message $Y=(y_1,...,y_m)$ as decoder input based on the NMT (Neural machine translation) model. \cite{jiang2017automatically, loyola2017neural, van2019generating}
However, taking the code modification without distinguishing between the added and the deleted part as model input makes it difficult to understand the context of modification in the NMT model.
In addition, previous studies tend to train from scratch when training a model, but this method does not show good performance because it creates a large gap in the contextual representation between programming language (PL) and natural language (NL).
To overcome the problems in previous studies and train a better commit message generation model, our approach follows two stages:

(1) Collecting and processing data with the pair of the added and deleted parts of the code $X=((add_1,del_1),...,(add_n,del_n))$.
To take this pair dataset into the Transformer-based NMT model \cite{vaswani2017attention}, we use the BERT \cite{devlin2018bert} fine-tuning method about two sentence-pair consist of added and deleted parts.
This shows a better BLEU-4 score \cite{papineni2002bleu} than previous works using raw git diff.
Similar to CodeSearchNet \cite{husain2019codesearchnet}, our data is also collected for six languages (Python, PHP, Go, Java, JavaScript, and Ruby) from Github to show good performance in various languages.
We finally released 345K code modification and commit message pair data.

(2) To solve a large gap about contextual representation between programming language (PL) and natural language (NL), we use CodeBERT \cite{feng2020codebert}, a language model well-trained in the code domain as the initial weight.
Using CodeBERT as the initial weight shows that the BLEU-4 score for commit message generation is better than when using random initialization and RoBERTa \cite{liu2019roberta}.
Additionally, when we pre-train the Code-to-NL task to document the source code in CodeSearchNet and use the initial weight of commit generation, the contextual representation between PL and NL is further reduced.

\section{Related Work}
Commit message generation has been studied in various ways.
\citet{jiang2017towards} collect 2M commits from the \citet{mauczka2015dataset} and top 1K Java projects in Github.
Among the commit messages, only those that keep the format of "Verb + Object" are filtered, grouped into verb types with similar characteristics, and then the classification model is trained with the naive Bayes classifier.

\citet{jiang2017automatically} use the commit data collected by \citet{jiang2017towards} to generate the commit message using an attention-based RNN encoder-decoder NMT model.
They filter again in a "verb/direct-object pattern" from 2M data and finally used the 26K commit message data.
\citet{loyola2017neural} uses an NMT model similar to \citet{jiang2017automatically}, but uses git diff and commit pairs collected from 1$\sim$3 repositories of Python, Java, JavaScript, and C++ as training data.
\citet{liu2018neural} propose a retrieval model using \citet{jiang2017automatically}'s 26K commit as training data.
Code modification is represented by bags of words vector, and the message with the highest cosine similarity is retrieved.
\citet{xu2019commit} collect only '.java' file format from \citet{jiang2017towards} and use 509K dataset as training data for NMT.
Also, to mitigate the problem of Out-of-Vocabulary (OOV) of code domain input, they use generation distribution or copying distribution similar to pointer-generator networks \cite{see2017get}.
\citet{van2019generating} also argues that the \citet{jiang2017towards} entire data is noise and proposes a pre-processing method that filters the better commit messages.

\citet{liu2020atom} argue that it is challenging to represent the information required for source code input in the NMT model with a fixed-length. In order to alleviate this, it is suggested that only the added and deleted parts of the code modification be abbreviated as abstract syntax tree (AST) and applied to the Bi-LSTM model.

\citet{niebcoregen} presented a large gap between the contextual representation between the source code and the natural language when generating commit messages.
Previous studies have used RNN or LSTM model, they use the transformer model, and similarly to other studies, they use \citet{liu2018neural} as the training data.
To reduce this gap, they try to reduce the two-loss that predict the next code line (Explicit Code Changes) and the randomly masked word in the binary file.
\section{Background}

\subsection{Git Process}
\label{sec:git_process}

Git is a version management system that manages version history and helps collaboration efficiently.
Git tracks all files in the project in the Working directory, Staging area, and Repository.
The working directory shows the files in their current state.
After modifying the file, developers move the files to the staging area using the \mintinline{shell}{add} command to record the modified contents and write a commit message through the \mintinline{shell}{commit} command.
Therefore, the commit message may contain two or more file changes.

\subsection{Text Summarization based on Encoder-Decoder Model}
\label{sec:enc_dec}

With the advent of sequence to sequence learning (Seq2Seq) \cite{sutskever2014sequence}, various tasks between the source and the target domain are being solved.
Text summarization is one of these tasks, showing good performance through the Seq2Seq model with a more advanced encoder and decoder.
The encoder and decoder models are trained by maximizing the conditional log-likelihood below based on source input $X=(x_1,...,x_n)$ and target input $Y=(y_1,...,y_m)$.
\begin{equation*}
\begin{aligned}
    p(Y|X;\theta)   &= log \sum_{t=0}^{T} p(y_t|y_{<t}, X;\theta)
\end{aligned}
\end{equation*}
where $T$ is the length of the target input, $y_0$ is the start token, $y_T$ is the end token and $\theta$ is the parameter of the model.

In the Transformer \cite{vaswani2017attention} model, the source input is vectorized into a hidden state through self-attention as the number of encoder layers.
After that, the target input also learns the generation distribution through self-attention and attention to the hidden state of the encoder.
It shows better summarization results than the existing RNN-based model \cite{nallapati2016abstractive}.

To improve performance, most machine translations use beam search.
It keeps the search area by $K$ most likely tokens at each step and searches the next step to generate better text.
Generation stops when the predicted $y_t$ is an end token or reaches the maximum target length.

\subsection{CodeSearchNet}
\label{sec:codesearchnet}
CodeSearchNet \cite{husain2019codesearchnet} is a dataset to search code function snippets in natural language.
It is a paired dataset of code function snippets for six programming languages (Python, PHP, Go, Java, JavaScript and Ruby) and a docstring summarizing these functions in natural language.
A total of 6M pair datasets is collected from projects with a re-distribution license.
Using the CodeSearchNet corpus, retrieval of the code corresponding to the query composed of natural language can be resolved.
Also, it is possible to resolve the problem of documenting the code by summarizing it in natural language (Code-to-NL).

\subsection{CodeBERT}
Recent NLP studies have shown state-of-the-art in various tasks through transfer learning consisting of pre-training and fine-tuning \cite{peters2018deep}.
In particular, BERT \cite{devlin2018bert} is a pre-trained language model by predicting masked words from randomly masked sequence input and uses only encoder based on Transformer \cite{vaswani2017attention}.
It shows good perfomances in various datasets and is now extending out of the natural language domain to the voice, video, and code domains.

CodeBERT is a pre-trained language model in the code domain to learn the relationship between programming language (PL) and natural language (NL).
In order to learn the representation between different domains, they refer to the learning method of ELECTRA \cite{clark2020electra} which is consists of Generator-Discriminator.
NL and Code Generator predict words from code tokens and comment tokens masked at a specific rate. Finally, NL-Code Discriminator is CodeBERT after trained through binary classification that predicts whether it is replaced or original.

CodeBERT shows good results for all tasks in the code domain.
Specially, it shows a higher score than other pre-trained models in the code to natural language(Code-to-NL) and code retrieval task from NL using CodeSearchNet Corpus.
In addition, CodeBERT uses the Byte Pair Encoding (BPE) tokenizer \cite{sennrich2015neural} used in RoBERTa, and does not generate unk tokens in code domain input.
\section{Dataset}
\label{sec:dataset}

We collect a 345K code modification dataset and commit message pairs from 52K repositories of six programming languages (Python, PHP, Go, Java, JavaScript, and Ruby) on Github.
When using raw git diff as model input, it is difficult to distinguish between added and deleted parts, so unlike \citet{jiang2017towards}, our dataset focuses only on the added and deleted lines in git diff.
The detailed data collection and pre-processing method are shown as a pseudo-code in Algorithm \ref{fig:pseudo}:

\begin{algorithm}
    \begin{algorithmic}[1]
    \caption{\label{fig:pseudo} Code modification parser from the list of repositories.}
    \Procedure{RepoParser($Repos$)}{}
    \For{$Repo$ in $Repos$}
        \State $commits = get\_commits(Repo)$
        \For{$commit$ in $commits$}
            \State $mods = get\_modifications(commit)$
            \For{$mod$ in $mods$}
                \If {$filtering(mod, commit)$}
                    \State \textbf{break}
                \EndIf
                \State Save $(mod.add, mod.del)$ to dataset.
            \EndFor
    	\EndFor
    \EndFor

    \EndProcedure
    \end{algorithmic}
\end{algorithm}
\begin{figure}[t]
    \centering 
    \includegraphics[width=0.44\textwidth]{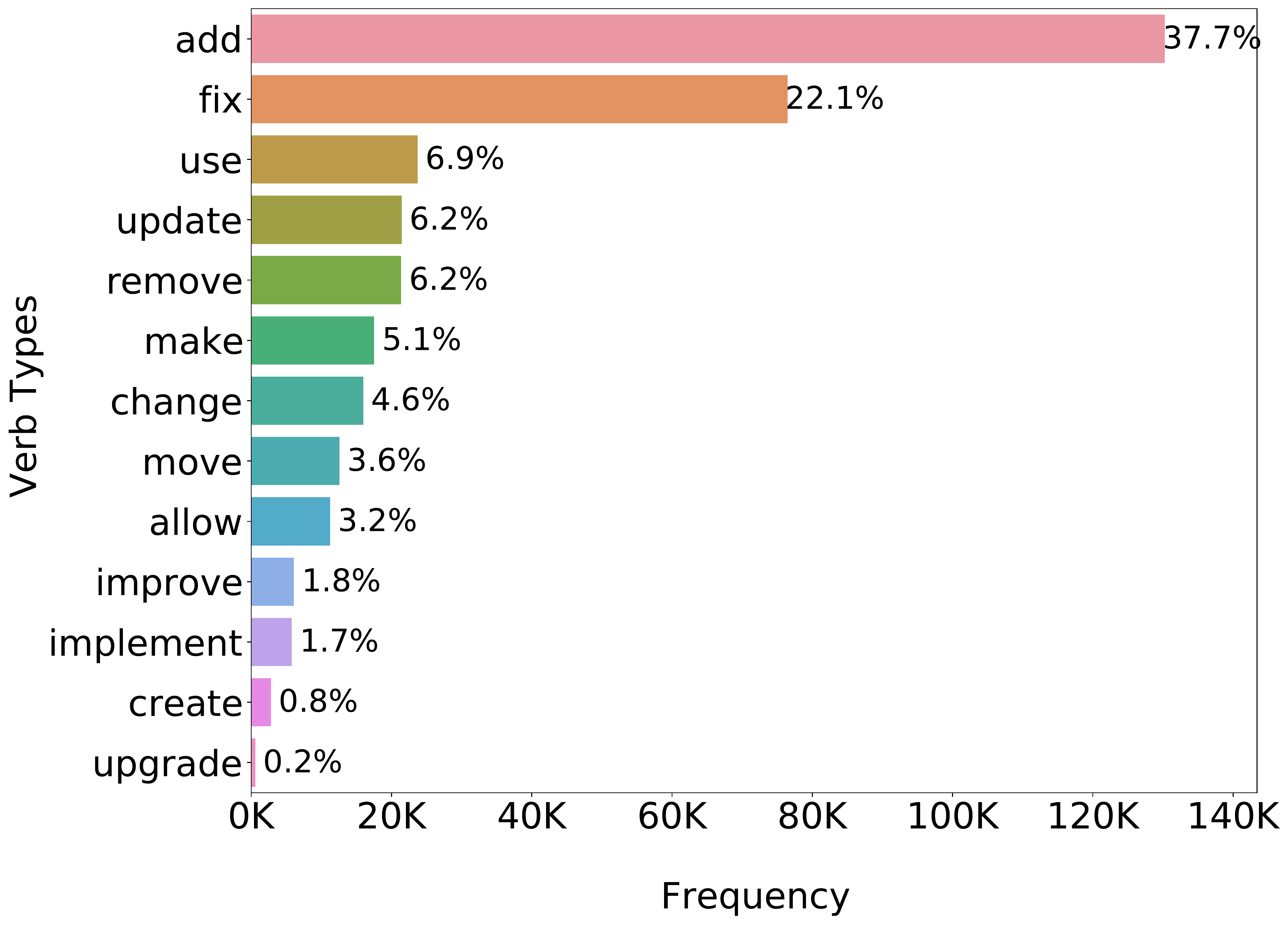}
    \caption
    {
    \label{fig:verb}
    Commit message verb type and frequency statistics.
    Only 'upgrade' is not included in the high frequency, but is included in a similar way to 'update'. This refers to the verb group in \citet{jiang2017towards}.
    } 
\end{figure}

To collect only the code that is a re-distributable license, we have listed the Github repository name in the CodeSearchNet dataset.
After that, all the repositories are cloned through multi-threading.
Detailed descriptions of functions that collect the commit hashes in a repository and the code modifications in a commit hash are as follows:

\begin{itemize}[noitemsep]
    \item \mintinline{python}{get_commits} is a function that gets the commit history from the repository.
    At this time, the commits of the master branch are filtered, excluding merge commits.
    Commits with code modifications corresponding to 6 the program language(.py, .php, .js, .java, .go, .ruby) extensions are collected. To implement this, we use the open-source pydriller \cite{spadini2018pydriller}.
    \item \mintinline{python}{get_modifications} is a function that gets the line modified in the commit.
    Through this function, it is possible to collect only the added or deleted parts, not all git diffs.
\end{itemize}

While collecting the pair dataset, we find that the relationship between some code modifications and the corresponding commit message is obscure and very abstract.
Also, we check that some code modification or commit message is a meaningless dummy file.
To filter these, we create the \mintinline{python}{filtering} function and the rules as follows.

\begin{enumerate}[noitemsep]
    \item To collect commit messages with various format distributions, we limit the collection of up to 50 commits in one repository.
    \item We filter commits whose number of files changed is one or two per commit message.
    \item Commit message with issue number is removed because detailed information is abbreviated.
    \item Similar to \citet{jiang2017towards}, the non-English commit messages are removed.
    \item Since some commit messages are very long, the first line is fetched.
    \item If the token of code through tree-sitter\footnote{\href{https://tree-sitter.github.io/tree-sitter}{https://tree-sitter.github.io/tree-sitter}}, a parser generator tool, exceeds 32 characters, it is excluded. This removes unnecessary things like changes to binary files in code diff.
    \item By referring to the \citet{jiang2017towards} and Conventional Commits(\S~\ref{sec:introduction}) rules, the commit message that begins with a verb is collected.
    We use spaCy\footnote{\href{https://spacy.io}{https://spacy.io}} for Pos tagging.
    \item We filter commit messages with 13 verb types, which are the most frequent.
    Figure \ref{fig:verb} shows the collected verb types and their ratio for the entire dataset.
\end{enumerate}

\begin{table}[t]
    \centering
    \footnotesize
    
    \begin{tabular*}{0.48\textwidth}{ccccc}
    \toprule
    \multicolumn{1}{c}{} & 
    \multicolumn{3}{c}{Number of Pair Dataset} \\
    & Train & Validation & Test & \makecell{Number of \\ Repositories} \\
    \midrule
    Python & 81517 & 10318 & 10258 & 12361 \\
    PHP  & 64458 & 8079 & 8100 & 16143  \\
    JavaScript  & 50561 & 6296 & 6252 & 11294  \\
    Ruby  & 29842 & 3772 & 3680 & 4581  \\
    Java  & 28069 & 3549 & 3552 & 4123  \\
    Go  & 21945 & 2699 & 2812 & 3960  \\
    \midrule
    \multicolumn{2}{l}{Total : 345759} & 
    \multicolumn{2}{c}{} &
    \multicolumn{1}{c}{52462} \\
    \bottomrule
    \end{tabular*}
    \caption{
    \label{tab:statistics}
    Dataset Statistics for each language collected from 52K repositories of six programming languages.
    }
\end{table}

As a result, we collect 345K code modification and commit message pair datasets from 52K Github repositories and split commit data into 80-10-10 train/validation/test sets.
This results are shown in Table \ref{tab:statistics}.
\section{CommitBERT}
\label{sec:commitbert}

We propose the idea of generating a commit message through the CodeBERT model with the our dataset (\S~\ref{sec:dataset}).
To this end, this section describes how to feed inputs code modification ($X=((add_1,del_1),...,(add_n,del_n))$) and commit message  ($Y=(msg1,...,msg_n)$) to CodeBERT and how to use pre-trained weights more efficiently to reduce the gap in contextual representation between programming language (PL) and natural language (NL).

\begin{figure*}[t!]
    \centering

    \subfloat[Code modification in git diff]{\includegraphics[width=0.27\textwidth]{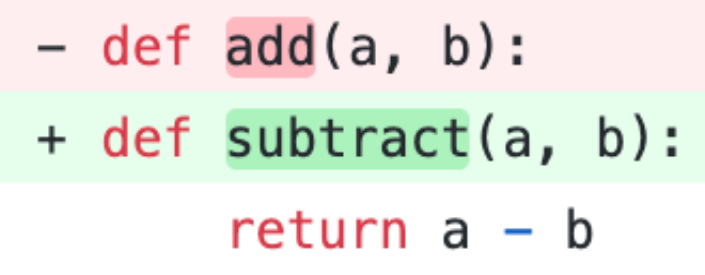}\label{fig:input_a}}
    \:\:\:\:
    \subfloat[CommitBERT input]{\includegraphics[width=0.65\textwidth]{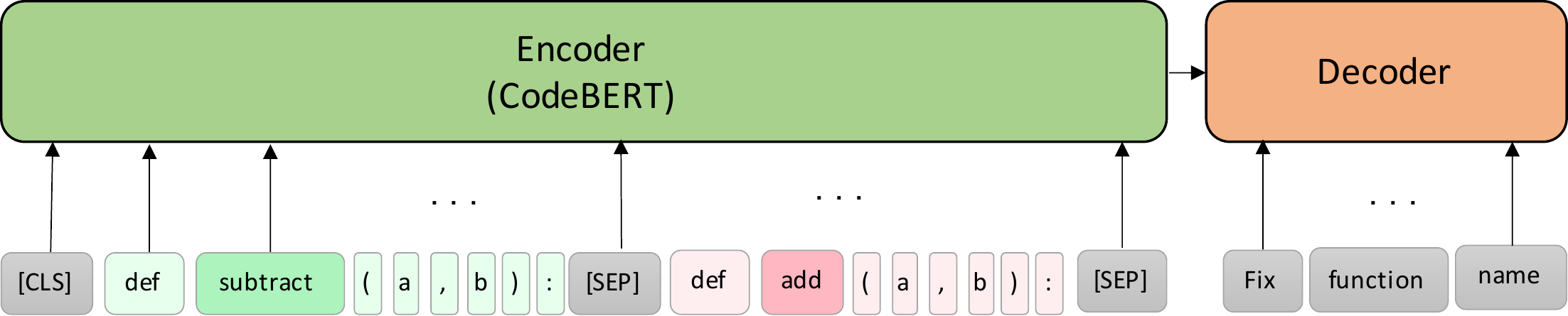}\label{fig:input_b}}

    \caption{
    \label{fig:commitbert}
    Illustration of a code modification example in git diff (a) and method of taking it to the input of CommitBERT (b).
    (b) shows that all code modification lines in (a) are not used, and only changed lines are as input.
    So, in this example, code modification (a) includes \mintinline{python}{return a - b}, but not in the model input (b).
    }
\end{figure*}

\subsection{CodeBERT for Commit Message Generation}
\label{sec:input}

We feed the code modification to the encoder and a commit message to the decoder input by following the NMT model.
Especially for code modification in the encoder, similar inputs are concatenated, and different types of inputs are separated by a sentence separator (sep).
Applying this to our CommitBERT in the same way, added tokens ($Add=(add_1,...,add_n)$) and deleted tokens ($Del=(del_1,...,del_n)$) of similar types are connected to each other, and sentence separators are inserted between them.
Therefore, the conditional-likelihood is as follows:

\begin{equation*}
\begin{aligned}
    p(M|C;\theta) &= log \sum_{t=0}^{T} p(m_t|m_{<t}, C;\theta), \\
    m_{<t} &= (m_0,m_1,...,m_{t-1}) \\
    C &= concat([cls], Add, [sep], Del, [sep]) \\
\end{aligned}
\end{equation*}

where $M$ is commit message tokens, $C$ is code modification tokens and $concat$ is list concatenation function.
$[cls]$ and $[sep]$ are speical tokens, which are a start token and a sentence separator token respectively.
Other notions are the same as Section \ref{sec:enc_dec}.

Unlike previous works, all code modifications in git diff are not used as input and only changed lines in code modification are used.
Since this removes unnecessary inputs, it shows a significant performance improvement in summarizing code modifications in natural language.
Figure \ref{fig:commitbert} shows how the code modification is actually taken as model input.

\subsection{Initialize Pretrained Weights}
\label{sec:init_weight}

To reduce the gap difference between two domains(PL, NL), We use the pretrained CodeBERT as the initial weight.
Furthermore, we determine that removing deleted tokens from our dataset (\S~\ref{sec:dataset}) is similar to the Code-to-NL task in CodeSearchNet (Section \ref{sec:codesearchnet}).
Using this feature, we use the initial weight after training the Code-to-NL task with CodeBERT as the initial weight.
This method of training shows better results than only using CodeBERT weight in commit message generation.
\section{Experiment}

\begin{table*}[t]
    \centering
    \footnotesize
    
    \begin{tabular*}{0.91\textwidth}{clcccccc}
    \toprule
    \multicolumn{1}{c}{Metric} & 
    \multicolumn{1}{c}{Initial Weight} & 
    \multicolumn{1}{c}{Python} & 
    \multicolumn{1}{c}{PHP} &
    \multicolumn{1}{c}{JavaScript} &
    \multicolumn{1}{c}{Java} & 
    \multicolumn{1}{c}{Go} &
    \multicolumn{1}{c}{Ruby} \\
    \midrule
    \multirow{4}*{BLEU-4 (Test)} & (a) Random & 7.95 & 7.01 &  8.41 & 7.60 & 10.38 & 7.17 \\
    & (b) RoBERTa & 10.94 & 9.71 & 9.50 & 6.40 & 10.21 & 8.95 \\
    & (c) CodeBERT & 12.05 & 13.06 & 10.47 & 8.91 & 11.19 & 10.33 \\
    & (d) CodeBERT + Code-to-NL & \textbf{12.93} & \textbf{14.30} & \textbf{11.49} & \textbf{9.81} & \textbf{12.76} & \textbf{10.56} \\
    \midrule[0.1pt]
    \multirow{4}*{PPL (Dev)} & (a) Random & 144.60 & 138.39 & 195.98 & 275.84 & 257.29 & 207.67 \\
    & (b) RoBERTa & 76.02 & 81.97 & 103.48 & 164.32 & 122.70 & 104.68 \\
    & (c) CodeBERT & 68.18 & 63.90 & 94.62 & 116.50 & 109.43 & 91.50 \\
    & (d) CodeBERT + Code-to-NL & \textbf{49.29} & \textbf{47.89} & \textbf{75.53} & \textbf{77.80} & \textbf{64.43} & \textbf{82.82} \\
    \bottomrule
    \end{tabular*}

    \caption{
    \label{tab:weight}
    Commit message generation result for 4 initial weights.
    In (c), CodeBERT is used as the initial weight. And (d) uses the weight trained on the Code-to-NL task in CodeSearchNet with CodeBERT as the initial weight.
    As a result, it shows BLEU-4 for the test set after training and the best PPL for the validation set in the during training.
    }
\end{table*}

\begin{table}[t]
    \centering
    \footnotesize
    
    \begin{tabular*}{0.48\textwidth}{clc}
    \toprule
    \multicolumn{1}{c}{\multirow{2}*{Initial Weight}} & 
    \multicolumn{1}{c}{\multirow{2}*{Input Type}} & 
    \multicolumn{1}{c}{\multirow{2}*{BLEU-4}} 
    \\ \\
    \midrule
    \multirow{2}*{RoBERTa} 
    & (a) All code modification  &  10.91 \\
    & (b) Only changed lines (\textbf{Ours}) &  \textbf{12.52} \\
    \midrule[0.1pt]
    \multirow{2}*{CodeBERT} 
    & (a) All code modification  &  11.77 \\
    & (b) Only changed lines (\textbf{Ours}) &  \textbf{13.32} \\
    \bottomrule
    \end{tabular*}

    \caption{
    \label{tab:input}
    The result of generating the commit message for the input type after collecting 4135 data with only source code change among the data of \citet{loyola2017neural}.
    (a) uses entire git diff(unidiff) as input, and (b) uses only the changed line according to Section \ref{sec:input} as input.
    }
\end{table}

To verify the proposal in Section \ref{sec:commitbert} in the commit message generation task, we do two experiments.
(1) Compare the commit message generation results of using all code modifications as inputs and using only the added or deleted lines as inputs.
(2) Ablation study several initial model weights to find the weight with the smallest gap in contextual representation between PL and NL.

\subsection{Experiment Setup}

Our implementation uses CodeXGLUE's code-text pipeline library \footnote{\href{https://github.com/microsoft/CodeXGLUE}{https://github.com/microsoft/CodeXGLUE}}.
We use the same model architecture and experimental parameters for the two experiments below.
As a model architecture, the encoder and decoder use 12 and 3 Transformer layers.
We use 5e-5 as the learning rate and train on one V100 GPU with a 32 batch size.
We also use 256 as the maximum source input length and 128 as the target input length, 10 training epochs, and 10 as the beam size $k$.

\subsection{Compare Model Input Type}
\label{sec:input_type}

To experiment generating a commit message according to the input type, only 4135 data is collected from data with code modification in the `.java' files among 26K training data of \citet{loyola2017neural}.
Then we transform these 4135 data into two types, respectively, and experiment with training data for RoBERTa and CodeBERT weights: (a) entire code modification in git diff and (b) only changed lines in code modification.
Figure \ref{fig:commitbert} shows these two differences in detail.

Table \ref{tab:input} shows the BLEU-4 values when inference with the test set after training about these two types.
Both initial weights show worse results than (b), even though type (a) takes a more extended input to the model.
This shows that lines other than changed lines as input data disturb training when generating the commit message.

\subsection{Ablation study on initial weight}

We do an ablation study while changing the initial weight of the model for 345K datasets in six programming languages collected in Section \ref{sec:dataset}.
As mentioned in \ref{sec:init_weight}, when the model weight with high comprehension in the code domain is used as the initial weight, it is assumed that the large gap in contextual representation between PL and NL would be greatly reduced.
To prove this, we train the commit message generation task for four weights as initial model weights:
Random, RoBERTa\footnote{\href{https://huggingface.co/roberta-base}{https://huggingface.co/roberta-base}}, CodeBERT\footnote{\href{https://huggingface.co/microsoft/codebert-base}{https://huggingface.co/microsoft/codebert-base}}, and the weights trained on the Code-to-NL task(Section \ref{sec:codesearchnet}) with CodeBERT.
Except for this initial weight, all training parameters are the same.

Table \ref{tab:weight} shows BLEU-4 for the test set and PPL for the dev set for each of the four weights after training.
As a result, using weights trained on the Code-to-NL task with CodeBERT as the initial weight shows the best results for test BLEU-4 and dev PPL.
It also shows good performance regardless of programming language.
\section{Conclusion and Future Work}

Our work presented a model summarizing code modifications to solve the difficulty of humans manually writing commit messages.
To this end, this paper proposed a method of collecting data, a method of taking it to a model, and a method of improving performance.
As a result, it showed a successful result in generating a commit message using our proposed methods.
Consequently, our work can help developers who have difficulty writing commit messages even with the application.

Although it is possible to generate a high-quality commit message with a pre-trained model, future studies to understand the code syntax structure remain in our work.
As a solution to this, CommitBERT should be converted to AST (Abstract Syntax Tree) before code modification is taken into the encoder like \cite{liu2020atom}.

\begin{table}[t]
    \centering
    \footnotesize
    \renewcommand{\arraystretch}{1.05} 
        
    \begin{tabular}{|c|p{5cm}|}
    \hline
    \makecell[c]{Language} & 
    \makecell[c]{Reference / Generated} \\
    \hline
    \multirow{2}*{Python} 
    & added figsize to plot methods \\
    \cline{2-2}
    & added figure size to plot\_weights \\
    \hline
    \multirow{2}*{PHP} 
    & added default value to fieldtype \\
    \cline{2-2}
    & Added default values of the fieldtype \\
    \hline
    \multirow{2}*{JavaScript} 
    & Fix missing = in delete uri \\
    \cline{2-2}
    & Fixed an issue with delete \\
    \hline
    \multirow{2}*{Java} 
    & Fixed the parsing of orders without a cid \\
    \cline{2-2}
    & Fix bug in exchange order \\
    \hline
    \multirow{2}*{Go} 
    & Use ioutil . Discard for benchmark \\
    \cline{2-2}
    & Use ioutil . Discard for logging \\
    \hline
    \multirow{2}*{Ruby} 
    & fixing schema validation issues with CCR export \\
    \cline{2-2}
    & fixing validation of ccr export \\
    \hline
    \end{tabular}
    
    \caption{
    \label{tab:result}
    The result of generating the commit message for six languages (Python, PHP, Go, Java, JavaScript, and Ruby) and the corresponding reference.
    We used the (d) model of Table \ref{tab:weight}.
    }
\end{table}

\section*{Acknowledgments}

The author would like to thank Gyuwan Kim, Dongjun Lee, Mansu Kim and the anonymous reviewers for their thoughtful paper review.

\bibliographystyle{acl_natbib}
\bibliography{anthology,acl2021}

\end{document}